\documentclass[letterpaper,10pt,conference]{ieeeconf}

\IEEEoverridecommandlockouts
\usepackage{amsmath,amssymb,amsfonts}
\usepackage{graphicx}
\usepackage{url}
\usepackage[table]{xcolor}
\usepackage[colorlinks=true,linkcolor=black,citecolor=black,urlcolor=blue!55!black]{hyperref}
\hypersetup{
  pdftitle={Refine Actions If Needed: Adaptive Action Chunk Refinement for Robot Manipulation},
  pdfauthor={Hanlin Liu, Zichen Zhou, Jiacheng Zhang, Liuyuan Na, Jin Jin, Guangyan Chen, Yufeng Yue, Gen Li, Fangqiang Ding}}

\providecommand\UseTaggingSocket[2]{}

\makeatletter
\providecommand{\bstctlcite}[1]{\@bsphack
  \@for\@citeb:=#1\do{%
    \edef\@citeb{\expandafter\@firstofone\@citeb}%
    \if@filesw\immediate\write\@auxout{\string\citation{\@citeb}}\fi}%
  \@esphack}
\makeatother

\newcommand{\tableouterrule}{\noalign{\hrule height 0.8pt}}
\newcommand{\tableheadrule}{\noalign{\hrule height 0.5pt}}
\newcommand{\tablegrouprule}{\noalign{\hrule height 0.3pt}}

\newif\ifblind
\blindfalse

\title{\LARGE\bfseries Refine Actions If Needed:\\Adaptive Action Chunk Refinement for Robot Manipulation}

\author{%
  \begin{minipage}{\textwidth}
  \centering
  {\normalsize Hanlin Liu\textsuperscript{1,2,*}\quad Zichen Zhou\textsuperscript{2,*}\quad
    Jiacheng Zhang\textsuperscript{1}\quad Liuyuan Na\textsuperscript{1,3}\endgraf}
  \vspace{3pt}
  {\normalsize Jin Jin\textsuperscript{4}\quad Guangyan Chen\textsuperscript{2}\quad
    Yufeng Yue\textsuperscript{2,\dag}\quad Gen Li\textsuperscript{5,\dag}\quad
    Fangqiang Ding\textsuperscript{1,\dag}\endgraf}
  \vspace{6pt}
  {\small
    \textsuperscript{1}The Hong Kong University of Science and Technology (Guangzhou)\endgraf
    \textsuperscript{2}Beijing Institute of Technology\quad \textsuperscript{3}Shanghai Jiao Tong University\endgraf
    \textsuperscript{4}University of Oxford\quad \textsuperscript{5}Nanyang Technological University\endgraf
    \vspace{3pt}
    \textsuperscript{*}Equal contribution.\quad \textsuperscript{\dag}Corresponding authors.\endgraf}
  \end{minipage}%
}

\begin{document}
\bstctlcite{IEEEauthorcontrol}

\flushbottom

\makeatletter
\IEEEaftertitletext{%
  \begin{minipage}{\textwidth}
    \def\@captype{figure}
    \centering
    \includegraphics[width=\textwidth]{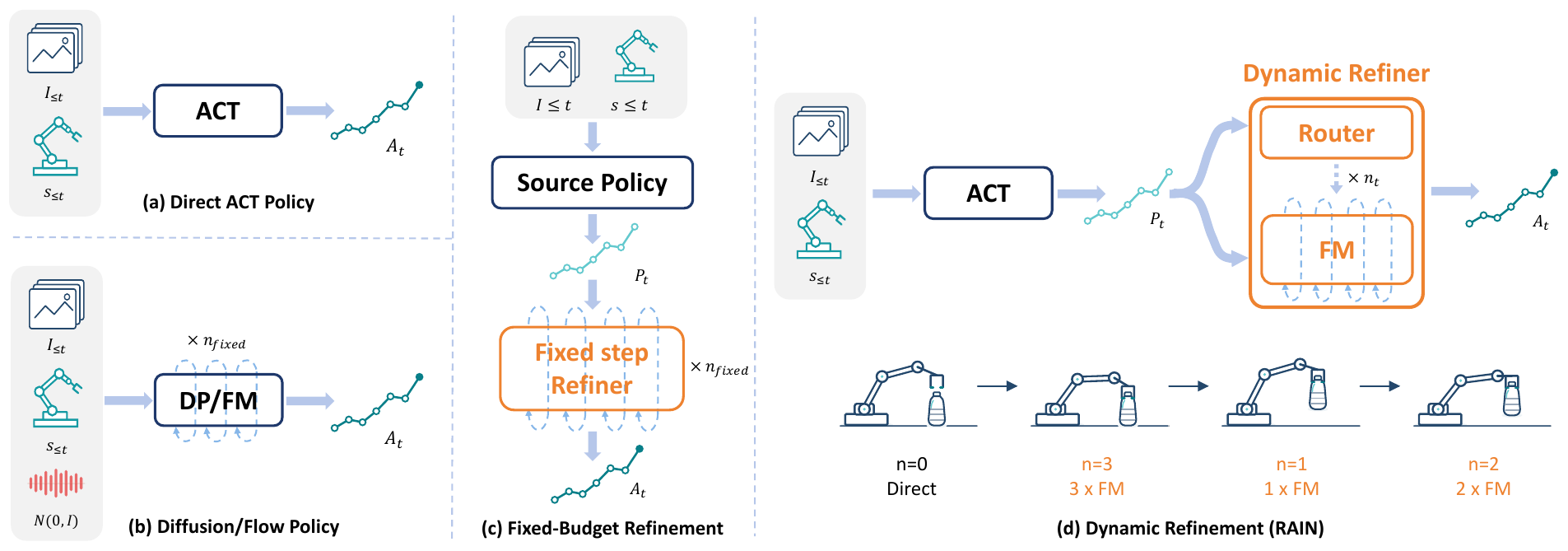}
    \caption{Comparison of action generation and refinement paradigms. (a) Direct ACT predicts an executable action chunk in a single forward pass. (b) Generative policies such as diffusion policy (DP) and flow matching (FM) generate action chunks through iterative network evaluations. (c) Fixed-budget refinement applies a fixed number of refinement steps to every action proposal. (d) RAIN predicts whether refinement is needed before invoking FM, directly executes proposals that require no refinement, and adaptively allocates additional refinement steps when necessary. The bottle-manipulation sequence illustrates how the computational budget varies across successive policy calls within a single episode.}
    \label{fig:paradigm-overview}
  \end{minipage}\vspace{\textfloatsep}%
}
\makeatother
\maketitle
\thispagestyle{plain}
\pagestyle{plain}

\begin{abstract}
Computationally lightweight manipulation policies predict executable actions with low inference latency, but prediction errors can limit task success. Iterative generative policies can achieve higher task success rates, but repeated network evaluations increase inference latency. Refining action proposals from computationally lightweight policies can combine these strengths, but the amount of refinement needed varies across states. We introduce RAIN (Refine Actions If Needed), a framework that predicts action errors at different refinement levels to decide whether and how much to refine each executable proposal, before performing any refinement. We train an ensemble to predict these errors by comparing action proposals and their refined versions with expert actions from offline demonstrations, without requiring explicit labels of refinement needs. Calibration balances uncertainty, computation, and temporal consistency. We evaluate RAIN on eight ManiSkill tasks and four real-robot tasks. In simulation, our policy achieves 66.6\% success, approaching the best evaluated diffusion-policy success rate with 5.71$\times$ lower policy latency. On the real robot, success increases from 55.0\% with Action Chunking with Transformers (ACT) to 72.5\%, with comparable average successful completion time. Matched controls show that dynamic routing maintains the strongest observed fixed-refinement success level with 26.6\% fewer refiner evaluations.

\par\smallskip
Project page: \url{https://rain-policy.github.io/}.
\end{abstract}

\section{Introduction}
\label{sec:introduction}

Closed-loop manipulation policies must generate effective actions quickly enough to replan as the environment changes. Diffusion Policy (DP) and flow-matching (FM) policies commonly generate these sequences by transforming an unstructured noise sample into an observation-conditioned action trajectory through repeated network evaluations~\cite{chi2023diffusion,lipman2023flow}. This iterative process can improve task success rates, but repeated network evaluations increase inference latency. To reduce inference latency, a computationally lightweight policy can first produce an executable action proposal for a generative model to refine. Instead of generating the entire action sequence from noise, the model can then focus on correcting errors in the proposal.

BRIDGeR and Action-to-Action (A2A) flow matching explore related approaches by replacing noise with policy samples or action-history representations~\cite{chen2024bridger,jia2026a2a}. However, improving the starting point does not by itself determine how much refinement is needed. For an executable proposal, the appropriate choice may range from direct execution to additional correction, depending on the current state. Figure~\ref{fig:paradigm-overview} contrasts direct generation, iterative generation, fixed refinement, and our policy with adaptive refinement. The problem is therefore to determine whether the expected improvement over direct execution justifies additional computation, before invoking the refiner.

Making this decision before refinement presents two challenges. First, predicting how much refinement will improve action quality requires assessing refined actions before they are generated. Proposal uncertainty alone does not establish whether the refiner can improve a proposal or how much correction is worthwhile, while evaluating alternative refinements online consumes the computation we seek to save. Second, offline demonstrations provide expert actions, but no direct labels for refinement needs or the benefit of each allocation. The router must therefore learn from demonstrations to predict action errors at different refinement levels, including no refinement. These predictions allow it to decide whether and how much to refine by weighing the expected reduction in error against the additional computation.

We introduce RAIN (Refine Actions If Needed) to address these challenges through offline comparative supervision and pre-refinement risk prediction. Here, risk refers to the predicted deviation of an action chunk from expert actions. At demonstration states, frozen proposal and refiner models generate candidates whose deviations from the same expert action chunk provide continuous surrogate targets for a risk ensemble. Offline calibration balances predicted risk, uncertainty, computation, and temporal consistency. The router then selects direct execution or only the chosen refinement, without online reward optimization or refinement probes. We instantiate this framework with Action Chunking with Transformers (ACT)~\cite{zhao2023act} as the proposal policy and flow matching as the refiner.

We evaluate the framework on eight ManiSkill tasks~\cite{tao2024maniskill3} and four real-robot tasks. In simulation, our policy achieves 66.6\% success at 0.322\,ms/action of amortized policy latency, approaching the best evaluated diffusion-policy success rate. On the real robot, it improves success over ACT by 17.5 percentage points with comparable average successful completion time. Matched controls further show that dynamic routing maintains the strongest observed fixed-refinement success level while reducing policy latency by 15.7\%.

The paper makes three contributions:
\begin{itemize}
  \item RAIN, an adaptive action-refinement framework that decides whether and how much to refine an executable action proposal before invoking a generative refiner, preserving a direct-execution path.
  \item A demonstration-only learning procedure that predicts continuous candidate risks from expert-action deviations, with offline calibration of uncertainty, computation cost, and temporal consistency.
  \item Evaluation on eight ManiSkill tasks and four real-robot tasks demonstrating the success--efficiency benefits of executable-proposal refinement, with matched controls isolating the additional computation savings from dynamic routing.
\end{itemize}

\section{Related Work}
\label{sec:related-work}

\begin{figure*}[t]
    \centering
    \includegraphics[width=0.96\textwidth]{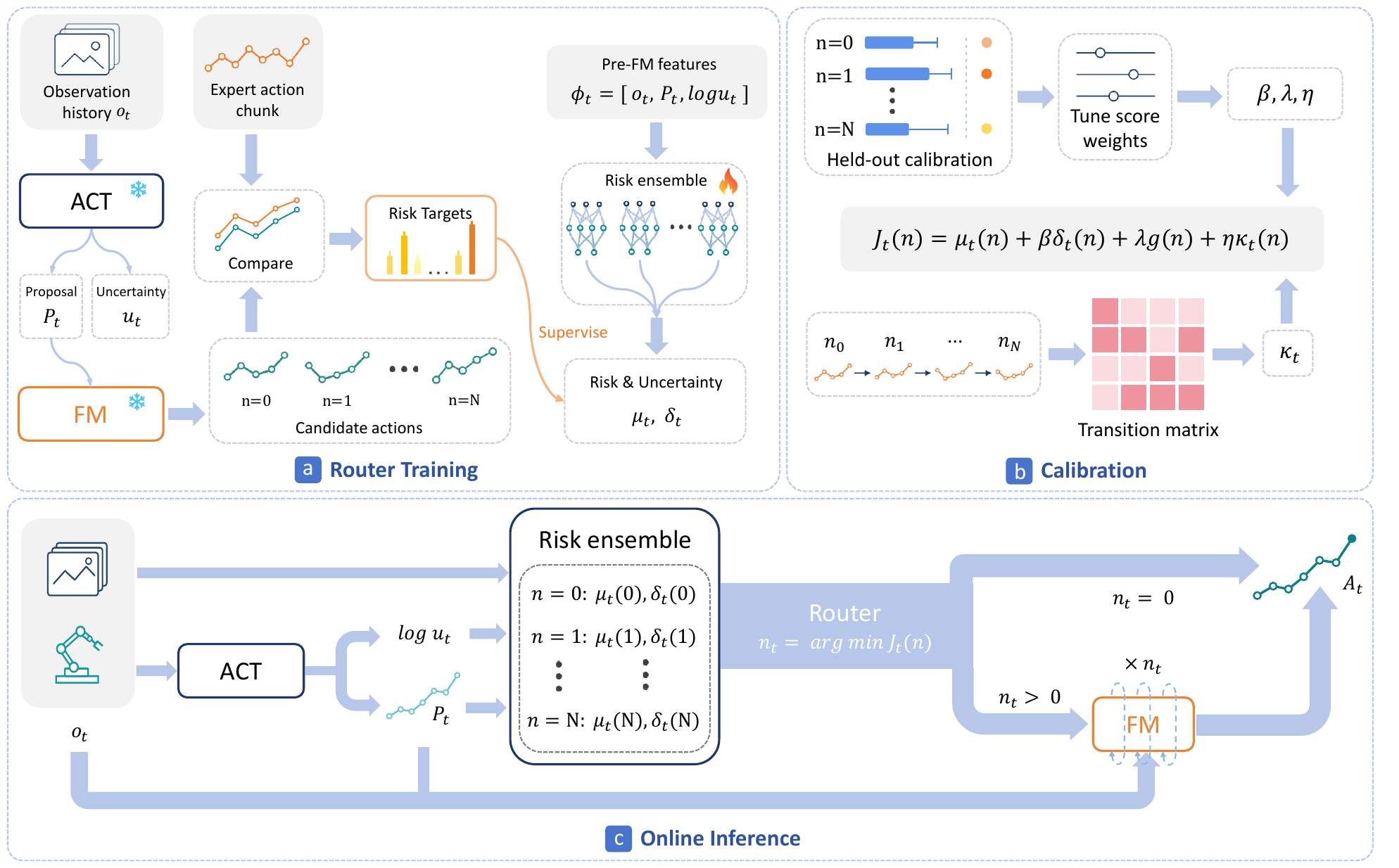}
    \caption{Overview of RAIN and its risk-routing mechanism. Expert-action deviations of unrefined and refined candidates supervise a risk ensemble; held-out trajectories calibrate uncertainty, compute, and temporal consistency; online, one pre-FM decision either executes the proposal or applies only the selected correction.}
    \label{fig:method-overview}
\end{figure*}

\textbf{Action policies and generative efficiency.}
Imitation policies model robot actions through direct prediction, implicit models, or tokenized representations. ACT predicts executable action chunks in one forward pass~\cite{zhao2023act}, while implicit behavior cloning and behavior transformers offer alternative representations of complex action distributions~\cite{florence2022ibc,shafiullah2022bet,lee2024vqbet}. Diffusion and flow-matching policies instead construct action sequences through iterative denoising or vector-field integration~\cite{chi2023diffusion,lipman2023flow,chisari2024pointflowmatch}. This expressive generation process requires repeated network evaluations at each control decision, motivating efforts to reduce inference cost.

Several approaches reduce this cost by shortening generation or reusing computation. Consistency Policy, One-Step Diffusion Policy, and FlowPolicy enable few-step or single-step prediction through distillation or consistency training~\cite{prasad2024consistency,wang2024onedp,zhang2024flowpolicy}. Warm-start methods exploit temporal continuity or predict informative initial actions~\cite{chen2025falcon,duan2025rti,li2026step}, while streaming generation and improved flow couplings offer complementary efficiency gains~\cite{hoeg2025streaming,jiang2025streamingflow,sochopoulos2025cot}. However, distillation can trade action accuracy or diversity for speed~\cite{hoeg2025streaming}, while temporal reuse alone may not ensure compatibility with the current state~\cite{li2026step}. These trade-offs motivate starting from an executable proposal and allocating correction according to its predicted value.

\textbf{Executable proposals and learned refinement.}
Informative starting points offer another route to efficient generation. BRIDGeR refines source-policy samples~\cite{chen2024bridger}, while A2A and Vision-to-Action (VITA) construct flow sources from action history and visual representations~\cite{jia2026a2a,gao2026vita}. These methods show that generation need not begin from unstructured noise, but an informative source alone does not determine whether further refinement justifies its cost. When the source is already executable, refinement becomes optional. RAIN explicitly compares the predicted value and cost of correction against direct execution, making the proposal both a refinement starting point and an executable alternative.

Learning corrections to an existing policy is also central to residual-policy methods. Residual reinforcement learning, FISH, and ResiP improve base policies through environment interaction~\cite{johannink2019residual,haldar2023fish,ankile2025resip}, while TRANSIC obtains corrective supervision from human interventions during deployment~\cite{jiang2024transic}. These approaches improve existing behavior but require additional interaction or intervention data for correction learning. RAIN instead learns refinement and its allocation from offline demonstrations, using candidate deviations from expert actions to supervise relative correction value. This supports adaptive refinement without collecting online rewards or human corrections.

\textbf{Adaptive computation for generative policies.}
Adaptive computation allocates different generation effort to different control decisions. AdaFlow adjusts integration using conditional variance linked to discretization error~\cite{hu2024adaflow}, while ProbeFlow estimates trajectory complexity from initial and lookahead velocity vectors~\cite{fang2026probeflow}. Prefix-Optimal Generative Policies (POGP) instead learn intermediate denoising values to stop when further generation offers little benefit~\cite{salla2026pogp}. These criteria adapt computation within generation but do not directly compare refinement against an independently executable proposal. RAIN makes this comparison before the first refiner evaluation, allowing direct execution when the predicted correction benefit does not justify its cost.

Other approaches learn computation allocation from observations. Dynamic Denoising Diffusion Policy (D3P) jointly trains a step adaptor and diffusion policy through reinforcement learning~\cite{yu2025d3p}, whereas RAIN learns refinement allocation from offline demonstrations. Difficulty-Aware Stochastic Interpolant Policy (DA-SIP) selects integration steps and solver configurations, while the Vision-Adaptive Diffusion Policy Framework (VADF) adjusts denoising schedules and execution lengths~\cite{chun2025dasip,yu2026vadf}. These methods adapt generation or execution; RAIN instead compares correction amounts for the same executable proposal, including no correction. Modality-Adaptive Robot Sampling (MARS) selectively invokes stochastic generation when behavioral diversity is needed~\cite{jia2026mars}. Yet even a single-modal action can require precise correction. RAIN therefore predicts continuous candidate risks, preserving correction differences that can be weighed against uncertainty and computation cost.

\section{Method}
\label{sec:method}

As shown in Fig.~\ref{fig:method-overview}, RAIN starts from an executable action proposal and treats generative
refinement as optional correction. We first train the proposal policy and
refiner, then freeze them. Using offline demonstrations, we compare actions
at different refinement levels, including no refinement, with expert actions
to obtain error targets. We train an ensemble to predict these errors before
running the refiner. At deployment, these predictions guide whether and how
much to refine the proposal, and only the selected refinement amount is executed.

\subsection{Executable Proposals and Optional Refinement}
\label{sec:method-overview}

At each planning step $t$, the proposal policy uses the available observation
history $\mathbf o_t$ to produce an executable action chunk
$\mathbf P_t\in\mathbb R^{H\times d_a}$ and a scalar uncertainty signal
$u_t>0$. Here, $H$ is the chunk length and $d_a$ is the action dimension. In
our ACT implementation, $\mathbf P_t$ is constructed deterministically from
mean action predictions, including temporal aggregation where applicable. The
uncertainty summarizes predicted arm-action variance; it informs routing but
is never sampled into the proposal.

The refinement amount $n$ specifies the number of refiner evaluations, with
$n=0$ retaining direct execution:
\begin{equation}
 \mathbf A_t^{(n)}=
 \begin{cases}
 \mathbf P_t, & n=0,\\
 \pi_{\mathrm{FM}}^{(n)}(\mathbf P_t\mid\mathbf o_t), & n>0.
 \end{cases}
 \label{eq:refinement-candidates}
\end{equation}
Flow matching starts from the proposal rather than noise. Positive refinement
amounts integrate the learned velocity field to correct the arm-action
trajectory, while retaining the proposal's gripper commands. All candidates
start from the same proposal and have the same action horizon. The router
decides which candidate to use from information available before the first
refiner evaluation.

\subsection{Joint Proposal--Refiner Training}
\label{sec:proposal-refiner-training}

The proposal should remain useful both for direct execution and as a starting
point for correction. We therefore retain its supervised action-prediction
objective while jointly optimizing the refiner.

For an aligned expert action chunk $\mathbf A_t^E$, flow matching~\cite{lipman2023flow} interpolates
between the proposal and expert actions. Using normalized arm-action components
in the following expression, its training objective is
\begin{equation}
 \begin{aligned}
 \mathbf X_\tau&=(1-\tau)\mathbf P_t+\tau\mathbf A_t^E,\\
 \mathcal L_{\mathrm{FM}}&=\mathbb E\!\Bigl[
 \bigl\|v_\psi(\mathbf X_\tau,\tau\mid\mathbf o_t,\mathbf P_t)\\
 &\hspace{4em}-(\mathbf A_t^E-\mathbf P_t)\bigr\|_{\mathrm{valid}}^2
 \Bigr],
 \end{aligned}
 \label{eq:fm-training}
\end{equation}
where $\tau\sim\mathcal U[0,1]$, and the masked squared error averages over
valid action entries. The velocity field thus learns how to move a proposal
toward the demonstrated action sequence.

The joint objective is
\begin{equation}
 \mathcal L_{\mathrm{joint}}
 =\frac{\mathcal L_{\mathrm{FM}}}{\bar L_{\mathrm{FM}}}
 +\alpha_{\mathrm{ACT}}
 \frac{\mathcal L_{\mathrm{ACT}}}{\bar L_{\mathrm{ACT}}},
 \label{eq:joint-loss}
\end{equation}
where the reference magnitudes are fixed before joint optimization. The ACT
loss trains the proposal policy to predict actions and their variance. The
flow-matching loss also updates the proposal policy, encouraging it to produce
action chunks that the refiner can improve. Joint training thus encourages
proposals that are both directly executable and easier to refine. Both models are subsequently frozen for
router training.

\subsection{Offline Candidate-Risk Supervision}
\label{sec:offline-risk}

To learn when correction is useful, we evaluate alternative refinement amounts
at demonstration states. Each candidate is compared with the same aligned
expert chunk:
\begin{equation}
 \begin{aligned}
 \ell_t(n)&=d_\gamma(\mathbf A_t^{(n)},\mathbf A_t^E),\\
 y_t(n)&=\log\!\left(1+\frac{\ell_t(n)}{\ell_{\mathrm{ref}}}\right).
 \end{aligned}
 \label{eq:risk-transform}
\end{equation}
Here, $d_\gamma$ is a masked, dimension-normalized arm-action error with
temporal weights $\gamma^h$, and $\ell_{\mathrm{ref}}>0$ is a shared
training-set scale. The logarithmic transformation reduces the influence of
large errors while preserving candidate ordering. All candidates, including
direct execution, receive continuous targets.

Expert-action deviation provides dense supervision for comparing alternatives
without branching the environment or collecting rollout rewards. A smaller
error means that the candidate action chunk is closer to the expert
demonstration. Comparing these errors across refinement levels helps determine
whether additional refinement improves the proposal.

\subsection{Pre-Refinement Risk Prediction}
\label{sec:risk-ensemble}

To predict both action errors and the uncertainty of these predictions, we use
an ensemble of $M$ independently initialized networks~\cite{lakshminarayanan2017ensembles}.
Each network receives observation features, the executable proposal, and $\log u_t$. Each member predicts a
mean and log variance for every candidate's transformed error. Training
combines heteroscedastic regression with a Huber term and auxiliary
pairwise-ranking and best-candidate objectives. The continuous targets retain
information about candidate differences instead of reducing supervision to a
single preferred refinement label.

After reversing target standardization, the predicted mean and log variance
on the transformed-error scale are $\hat y_t^{(k)}(n)$ and
$\hat\rho_t^{(k)}(n)$, respectively. We aggregate them as
\begin{equation}
 \begin{aligned}
 \mu_t(n)&=\max\!\left(0,\mathbb E_k[\hat y_t^{(k)}(n)]\right),\\
 \delta_t(n)&=\sqrt{\mathbb E_k[e^{\hat\rho_t^{(k)}(n)}]
 +\operatorname{Var}_k[\hat y_t^{(k)}(n)]}.
 \end{aligned}
 \label{eq:deployment-risk}
\end{equation}
Thus, $\mu_t(n)$ estimates transformed candidate error, while $\delta_t(n)$ combines
predicted variance and ensemble disagreement. Both are learned for direct
execution as well as refined candidates. Proposal uncertainty $u_t$ is an
input to this prediction; it is distinct from the ensemble's uncertainty
$\delta_t(n)$ and does not impose a separate refinement threshold.

\begin{figure*}[t]
    \vspace*{5pt}
    \centering
    \includegraphics[width=\textwidth]{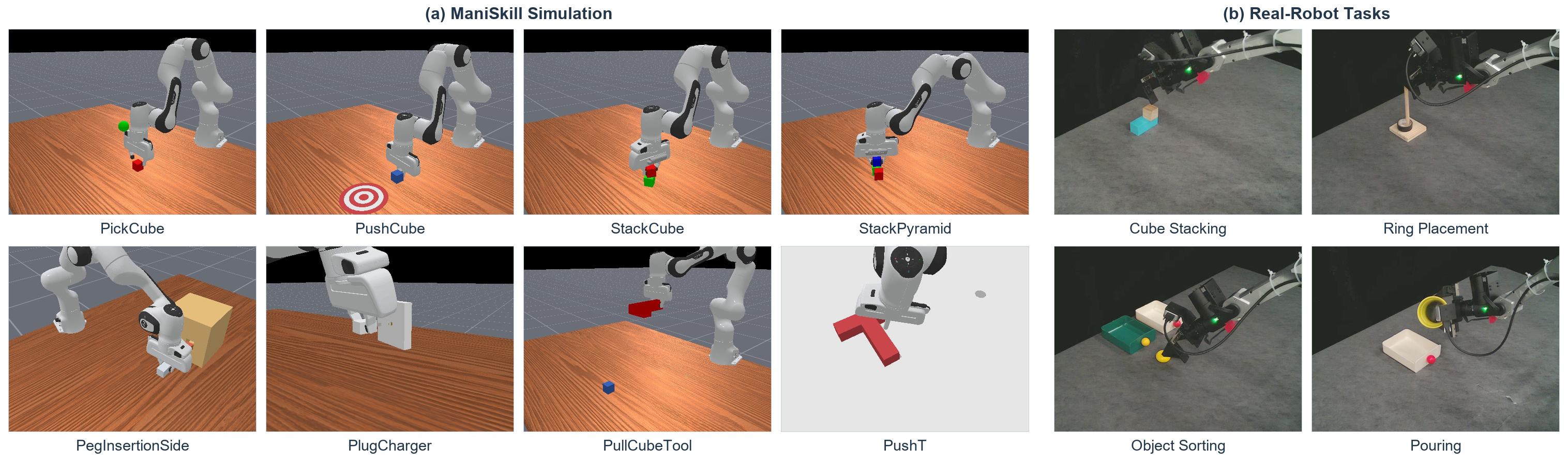}
    \caption{Task overview. (a) Eight ManiSkill3 manipulation tasks. (b) Four real-robot tasks: Cube Stacking, Ring Placement, Object Sorting, and Pouring. Real-robot images have been brightness-adjusted for visibility.}
    \label{fig:task-overview}
\end{figure*}

\subsection{Offline Calibration and Online Routing}
\label{sec:pre-route}

At deployment, the router balances predicted error against uncertainty,
computation, and temporal consistency:
\begin{equation}
 \begin{aligned}
 J_t(n)&=\mu_t(n)+\beta\delta_t(n)+\lambda g(n)+\eta\kappa_t(n),\\
 n_t&=\arg\min_n J_t(n).
 \end{aligned}
 \label{eq:routing-score}
\end{equation}
The nondecreasing cost $g(n)$, with $g(0)=0$, represents relative refinement
computation. A refined candidate is selected only when its combined score is
preferable to direct execution; a larger refinement amount is not assumed to
be better.

The temporal term captures how the refinement amount with the lowest
expert-action error changes along offline demonstration trajectories,
$n_t^\star=\arg\min_n\ell_t(n)$. We count transitions between these amounts
to estimate a smoothed transition probability matrix $\mathbf T$.
Let $n_{t^-}$ be the refinement amount chosen at the previous planning step,
and let $\Delta_t$ be the number of environment steps since that decision. Then,
\begin{equation}
 \kappa_t(n)=-\log\!\left(
 \max\!\left\{[\mathbf T^{\Delta_t}]_{n_{t^-},n},\epsilon\right\}
 \right).
 \label{eq:transition-cost}
\end{equation}
The first decision has zero transition cost. This soft prior favors plausible
refinement sequences without excluding any candidate.

We calibrate $\beta,\lambda,\eta$ on offline demonstration trajectories,
with the data allocation specified in Sec.~\ref{sec:experiments}. Settings
are evaluated by their trajectory-averaged disagreement with the minimum-error
candidate. Among settings whose excess error over the best setting is within
one paired standard error, we select the one with the lowest mean refinement
amount. The coefficients are then frozen before closed-loop evaluation.

Online, the ensemble predicts all candidate scores without running the
refiner. The policy either executes $\mathbf P_t$ directly or constructs only
the selected refined candidate. Alternative candidates are generated for
offline supervision, not for online search.

\section{Experiments}
\label{sec:experiments}

\begin{table*}[t]
\centering
\footnotesize
\setlength{\tabcolsep}{4.8pt}
\renewcommand{\arraystretch}{1.08}
\caption{Success rate and policy latency on eight ManiSkill tasks. Bold indicates the best result in each column. Rank 1 is best; ties share a rank, with subsequent ranks skipped.}
\label{tab:system-baselines}
\begin{tabular}{lrrrrrrrrcccc}
\tableouterrule
Method & Pick & Push & Stack & Peg & Plug & Pull & PushT & Pyramid & Overall success (\%) $\uparrow$ & Rank & Policy latency (ms/action) $\downarrow$ & Rank \\
\tableheadrule
ACT & 98 & 95 & 5 & 7 & 4 & 84 & 3 & 35 & 41.375 & 13 & 0.222 & 3 \\
\tablegrouprule
A2A-1 & 99 & 50 & 86 & 1 & 0 & 11 & 12 & 3 & 32.750 & 14 & \textbf{0.127} & \textbf{1} \\
A2A-4 & 99 & \textbf{100} & 90 & 8 & 10 & 31 & \textbf{29} & 36 & 50.375 & 11 & 0.206 & 2 \\
A2A-6 & \textbf{100} & \textbf{100} & 94 & 27 & 7 & 33 & \textbf{29} & 53 & 55.375 & 10 & 0.260 & 4 \\
\tablegrouprule
DP-1 & 0 & 0 & 0 & 0 & 0 & 4 & 0 & 0 & 0.500 & 15 & 0.378 & 7 \\
DP-2 & \textbf{100} & 99 & 95 & 40 & 7 & 74 & 21 & 32 & 58.500 & 8 & 0.746 & 8 \\
DP-5 & \textbf{100} & \textbf{100} & \textbf{98} & 48 & \textbf{22} & \textbf{99} & 20 & 56 & \textbf{67.875} & \textbf{1} & 1.839 & 12 \\
DP-10 & \textbf{100} & \textbf{100} & 97 & 56 & 14 & \textbf{99} & 15 & 59 & 67.500 & 2 & 3.120 & 14 \\
DP-100 & \textbf{100} & 99 & \textbf{98} & 55 & 17 & 98 & 21 & 52 & 67.500 & 2 & 38.517 & 15 \\
\tablegrouprule
FM-1 & \textbf{100} & 98 & 85 & 12 & 3 & 25 & 16 & 22 & 45.125 & 12 & 0.376 & 6 \\
FM-2 & \textbf{100} & \textbf{100} & 95 & 32 & 2 & 76 & 17 & 42 & 58.000 & 9 & 0.747 & 9 \\
FM-5 & \textbf{100} & \textbf{100} & 94 & 42 & 5 & 93 & 17 & 50 & 62.625 & 6 & 1.835 & 11 \\
FM-10 & \textbf{100} & \textbf{100} & \textbf{98} & 44 & 9 & 96 & 15 & 51 & 64.125 & 5 & 3.074 & 13 \\
ProbeFlow & \textbf{100} & 99 & 96 & 37 & 3 & 78 & 16 & 49 & 59.750 & 7 & 0.931 & 10 \\
\tablegrouprule
\rowcolor{orange!8}
\textbf{RAIN} & \textbf{100} & 96 & 91 & \textbf{59} & 11 & 89 & 24 & \textbf{63} & 66.625 & 4 & 0.322 & 5 \\
\tableouterrule
\end{tabular}
\end{table*}

We evaluate overall performance, dynamic versus fixed refinement, risk prediction versus direct refinement prediction, the transition prior, and task-level refinement allocation.

\subsection{Experimental Setup}

\textbf{Tasks and data.}
We evaluate on eight ManiSkill3 tasks~\cite{tao2024maniskill3} and four real-robot tasks shown in Fig.~\ref{fig:task-overview}. Simulation policies use state observations, while real-robot policies use two RGB camera views and proprioceptive observations. Each simulated task provides 500 successful demonstrations. All simulation baselines are trained on all 500 demonstrations, while our framework uses 400 for training, 50 for calibration, and 50 held out from fitting and selection. Real-robot experiments use a six-degree-of-freedom ARX arm with a gripper. Each real task provides 50 training demonstrations recorded at 30 Hz. Each method is evaluated on 100 shared environment seeds per simulated task and 20 trials per real task.

\textbf{Compared methods.}
We compare RAIN with ACT, A2A, DP, FM, and ProbeFlow in simulation, and with ACT and DP on the robot. Numeric suffixes denote iterative network evaluations. We show A2A with 1, 4, and 6 evaluations. The simulation ACT baseline uses the official ACT checkpoint with an eight-action execution horizon. For ProbeFlow, we apply its adaptive solver to the frozen FM baseline without retraining, retaining Gaussian initialization. We use a probe time of 0.5, a score threshold of 0.008, and integration schedules of 2, 4, 6, 8, or 10 steps; latency includes both probe evaluations. For the matched simulation study, Fixed 0--3, Direct-C, and RAIN share frozen proposal/refiner checkpoints, preprocessing, evaluation seeds, and an eight-action execution block. Fixed $k$ applies $k$ refinements at each policy call, whereas RAIN selects the amount dynamically. Fixed 0 directly executes our jointly trained proposal, rather than the official ACT checkpoint. Simulation routers are fit and calibrated on the simulation splits. Real-robot router fitting and calibration reuse all 50 training demonstrations per task, without a separate calibration split.

\textbf{Metrics.}
We report per-task and overall success rates separately for simulation and real-robot experiments. Overall success is calculated across 800 simulated episodes or 80 real trials per method. Simulated episodes succeed when the environment first reports completion; real trials succeed when the task is completed. Simulation policy latency measures isolated 32-bit floating-point (FP32) CUDA-Graph inference on an otherwise idle NVIDIA A800, from observation availability to action-chunk readiness, excluding environment execution and host-to-device transfer. Latency and FM evaluations are reported per executed action. Real-robot task completion time is measured with policy inference on an NVIDIA RTX 4080 and includes observation processing, policy inference, and physical execution. We average completion time over successful trials within each task and then equally across tasks; unsuccessful trials contribute only to success rate.

\subsection{Overall Success Rate and Efficiency}

\noindent\textbf{Q1.} Can proposal refinement combine high success rates with efficient execution?
\par\nopagebreak[4]

\begin{figure}[!t]
    \centering
    \includegraphics[width=\columnwidth,trim=0 5bp 0 0,clip]{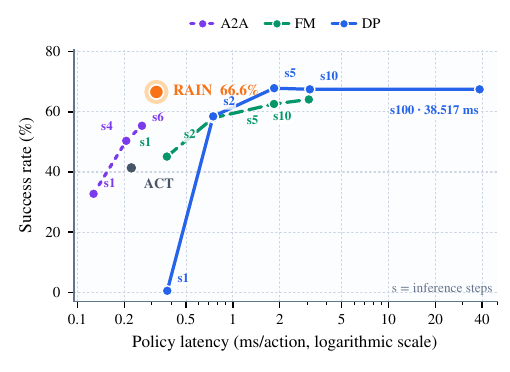}
    \caption{Success rate versus policy latency on a logarithmic latency axis. Colors and markers identify method families; numeric labels denote the number of network evaluations in each setting.}
    \label{fig:compute-success}
\par\medskip
\begingroup
\makeatletter\def\@captype{table}\makeatother
\centering
\footnotesize
\setlength{\tabcolsep}{4pt}
\renewcommand{\arraystretch}{1.3}
\caption{Success rate and task completion time on four real-robot tasks.}
\label{tab:real-robot}
\begin{tabular}{lcccccc}
\tableouterrule
\raisebox{-6pt}[0pt][0pt]{Method} & \multicolumn{5}{c}{\strut Success Rate (\%) $\uparrow$} & \raisebox{-6pt}[0pt][0pt]{Time (s) $\downarrow$} \\
\cline{2-6}
 & Stack & Ring & Sort & Pour & Overall & \\
\tableheadrule
ACT & 65 & 50 & 40 & 65 & 55 & \textbf{17.17} \\
DP & 70 & 50 & 55 & \textbf{70} & 61.25 & 21.49 \\
\rowcolor{orange!8}
\textbf{RAIN} & \textbf{85} & \textbf{65} & \textbf{70} & \textbf{70} & \textbf{72.5} & 17.57 \\
\tableouterrule
\end{tabular}
\par\smallskip
\parbox{\columnwidth}{\scriptsize Stack: Cube Stacking; Ring: Ring Placement; Sort: Object Sorting; Pour: Pouring. Time is averaged over successful trials within each task, then equally across tasks.}
\endgroup
\end{figure}

\textbf{ManiSkill results.} Table~\ref{tab:system-baselines} and Fig.~\ref{fig:compute-success} show that RAIN achieves a 66.6\% success rate at 0.322\,ms/action, the highest success rate among all evaluated methods below 1\,ms/action. We examine this operating point from two perspectives: the success gained over low-cost policies and the computation saved relative to iterative generative policies.

Compared with ACT, RAIN improves success from 41.4\% to 66.6\% with approximately 0.100\,ms/action of additional policy latency. Compared with the highest-success A2A setting, which generates actions from an action-history representation, RAIN gains 11.3 percentage points for only 0.062\,ms/action of additional latency. These comparisons show that our proposal-refinement system achieves substantially higher success while remaining close to these policies in inference cost.

Relative to iterative generative policies, the main advantage is retaining high success with substantially less computation. RAIN surpasses the FM setting with the highest success rate by 2.5 percentage points and reduces policy latency from 3.074 to 0.322\,ms/action, a 9.55$\times$ reduction. It also approaches the highest-success DP setting, which achieves 67.9\%, while reducing latency from 1.839 to 0.322\,ms/action, a 5.71$\times$ reduction. This advantage extends to the adaptive ProbeFlow baseline: RAIN raises success from 59.8\% to 66.6\% while reducing latency from 0.931 to 0.322\,ms/action, a 2.89$\times$ reduction. Thus, the evaluated proposal-refinement system improves both success and latency over the strongest FM setting and ProbeFlow, while approaching the best DP success rate at substantially lower latency.

\textbf{Real-robot results.} We next examine whether these benefits carry over to real-robot manipulation, where efficiency is measured by successful task completion time rather than isolated policy latency. RAIN achieves the highest overall success rate across the four tasks, reaching 72.5\%, compared with 55.0\% for ACT and 61.25\% for DP (Table~\ref{tab:real-robot}). It outperforms ACT on every task and matches or exceeds DP's success rate. This higher success is accompanied by an average successful completion time close to ACT's (17.57 versus 17.17\,s) and shorter than DP's (21.49\,s). The real-robot results therefore extend the simulation findings: executable-proposal refinement supports higher task success without a substantial increase in average successful completion time relative to ACT.

These comparisons establish the performance of the complete system; the following matched controls isolate the additional contribution of dynamic routing.

\subsection{Matched Routing Controls}

\noindent\textbf{Q2.} Can dynamic refinement improve success while reducing inference cost relative to fixed refinement?
\par\nopagebreak[4]

\begin{table}[!t]
\vspace*{5pt}
\centering
\footnotesize
\setlength{\tabcolsep}{2pt}
\renewcommand{\arraystretch}{1.04}
\caption{Success rate and inference cost of matched routing variants.}
\label{tab:routing-controls}
\begin{tabular}{lccc}
\tableouterrule
\raisebox{4pt}{Method} & \shortstack{Success\\(\%) $\uparrow$} & \shortstack{Latency\\(ms/action) $\downarrow$} & \shortstack{FM evals.\\(per action) $\downarrow$} \\
\tableheadrule
\rowcolor{black!6}
\multicolumn{4}{l}{Fixed allocation} \\
Fixed 0 & 46.5 & \textbf{0.158} & \textbf{0.000} \\
Fixed 1 & 65.1 & 0.278 & 0.125 \\
Fixed 2 & 65.8 & 0.330 & 0.251 \\
Fixed 3 & 66.4 & 0.382 & 0.376 \\
\rowcolor{black!6}
\multicolumn{4}{l}{Direct refinement prediction} \\
Direct-C Calibrated & 65.1 & 0.310 & 0.265 \\
Direct-C Argmax & 65.9 & 0.362 & 0.335 \\
\rowcolor{black!6}
\multicolumn{4}{l}{Transition-prior ablation} \\
RAIN w/o Transition Prior & 66.4 & 0.325 & 0.288 \\
\tablegrouprule
\rowcolor{orange!8}
\textbf{RAIN} & \textbf{66.6} & 0.322 & 0.276 \\
\tableouterrule
\end{tabular}
\end{table}

To isolate refinement allocation itself, we compare RAIN with Fixed 0--3 under the matched protocol in Table~\ref{tab:routing-controls}. All variants share the frozen proposal and refiner, preprocessing, evaluation seeds, eight-action execution horizon, and accelerated inference implementation; only the allocation of FM refinement across policy calls differs. This comparison separates the benefit of refinement from the additional benefit of deciding when and how much to refine.

Fixed 0 also outperforms the standalone ACT baseline, possibly because the additional variance-prediction objective helps the proposal policy learn task-relevant action variability.

Dynamic allocation achieves a slightly higher success rate with lower policy latency than the strongest fixed-refinement setting. RAIN reaches 66.6\% success at 0.322\,ms/action, compared with 66.4\% at 0.382\,ms/action for Fixed 3, reducing latency by 15.7\% and FM evaluations per action by 26.6\%. These results demonstrate a better success--latency operating point through state-dependent refinement.

RAIN also improves over Fixed 2, increasing success from 65.8\% to 66.6\% while reducing latency from 0.330 to 0.322\,ms/action, despite using slightly more FM evaluations.

\begin{figure}[t]
    \vspace*{3pt}
    \centering
    \includegraphics[width=\columnwidth]{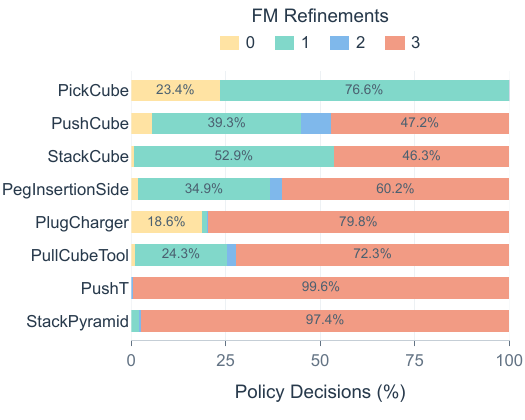}
    \caption{Task-level refinement allocation by RAIN. Each stacked bar shows the percentage of policy decisions using each refinement amount in our ManiSkill implementation, with each bar totaling 100\%.}
    \label{fig:routing-heatmap}
\end{figure}

\subsection{Continuous Risk versus Direct Refinement Prediction}

\noindent\textbf{Q3.} Does predicting candidate risks improve the success--compute trade-off over directly predicting refinement?
\par\nopagebreak[4]

To isolate the router supervision target, we compare RAIN with two direct refinement predictors under the matched protocol in Table~\ref{tab:routing-controls}. All variants use identical pre-refinement inputs and three-member ensembles. Direct-C converts each offline state into a hard label identifying the refinement amount with the lowest expert-action deviation, whereas RAIN predicts a numerical action-error estimate for each refinement level, referred to here as continuous risk. Direct-C Argmax deploys the raw class prediction, while Direct-C Calibrated applies offline uncertainty, compute, and transition calibration.

Calibration improves the efficiency of direct prediction, reducing latency from 0.362 to 0.310\,ms/action and FM evaluations per action from 0.335 to 0.265, with success changing from 65.9\% to 65.1\%. Continuous risk prediction offers a stronger success-oriented operating point: RAIN achieves 66.6\% success at 0.322\,ms/action, improving success while reducing both latency and refinement evaluations relative to Argmax. Compared with calibrated Direct-C, RAIN gains 1.5 percentage points of success for only 0.012\,ms/action of additional latency.

These results support predicting action errors for each refinement level rather than only predicting which level to choose. A single label does not distinguish between cases where refinement levels have similar errors and those where one has a much lower error. Predicting the error at each level preserves these differences, allowing the router to weigh the expected reduction in action error against the additional computation.

\subsection{Effect of the Transition Prior}

\noindent\textbf{Q4.} Does the transition prior improve refinement efficiency?
\par\nopagebreak[4]

With the proposal, refiner, and risk predictions held fixed, adding the transition prior reduces FM evaluations per action from 0.288 to 0.276 and policy latency from 0.325 to 0.322\,ms/action, while maintaining a comparable success rate of 66.6\% versus 66.4\% without the prior.

The prior complements current-state risk estimates with refinement-transition patterns learned from offline demonstrations. It favors plausible refinement sequences without excluding any candidate, allowing the current observation and proposal to continue determining the allocation. The ablation supports using this temporal information to further improve refinement efficiency, complementing the main allocation capability provided by candidate-risk prediction.

\subsection{How the Router Allocates Computation}

\noindent\textbf{Q5.} How does RAIN allocate refinement across tasks?
\par\nopagebreak[4]

Figure~\ref{fig:routing-heatmap} reveals distinct refinement profiles across tasks. PickCube relies almost entirely on zero or one refinement, while PushCube, StackCube, and PegInsertionSide use broader mixtures. PushT and StackPyramid, in contrast, use the highest refinement amount in 99.6\% and 97.4\% of decisions, respectively. The router therefore produces selective rather than uniformly reduced computation: it retains high refinement on some tasks while exploiting direct execution and lighter correction on others.

Together with the matched controls, these distributions explain how RAIN reduces computation without imposing a lower fixed refinement amount everywhere. Fixed 3 incurs the highest refinement cost at every decision, whereas RAIN selectively avoids that cost while maintaining a comparable overall success rate. The value of dynamic routing lies in this selective allocation, not in forcing balanced use of all refinement amounts or frequent switching.

\subsection{Refinement Beyond Flow Matching}

The main experiments instantiate generative refinement with flow matching. To examine whether executable proposals can also support diffusion-based correction, we consider an auxiliary ACT--diffusion experiment. Within this separate setting, refinement improves success from 52.9\% for the unrefined proposal to 63.6\% at the best evaluated fixed-refinement setting. These results provide supporting evidence that the proposal-refinement principle extends beyond flow matching; the dynamic routing results in this paper remain evaluated with the flow-matching refiner.

\section{Conclusion and Limitations}
\label{sec:conclusion}

\textbf{Conclusion.} We presented RAIN, a framework for adaptive generative refinement of executable action proposals. A computationally lightweight policy provides an action chunk, while a router learned from offline demonstrations decides whether and how much to refine it before invoking the generative model. Experiments on eight ManiSkill tasks and four real-robot tasks demonstrate high success rates at low simulation policy latency and improved real-robot success with successful task completion times comparable to ACT. Matched simulation controls further show that dynamic routing achieves a slightly higher success rate with lower latency than the strongest fixed-refinement setting. By starting from executable action proposals and dynamically adjusting refinement to the current environment, RAIN achieves high task success rates while avoiding unnecessary generative computation.

\textbf{Limitations.} The router learns from action errors relative to expert demonstrations, which may not fully reflect the effect of refinement on task success, especially in states outside the demonstration distribution. Learning refinement decisions that better predict task outcomes remains an important direction.

\let\originalthebibliography\thebibliography
\renewcommand{\thebibliography}[1]{\originalthebibliography{#1}\small\interlinepenalty=10000}
\IEEEtriggeratref{29}
\bibliographystyle{IEEEtran}
\bibliography{references}

\end{document}